\documentclass{llncs} %
\usepackage{float}
\usepackage{graphicx}
\usepackage{wrapfig}
\usepackage{placeins}
\usepackage[labelfont=bf]{caption}
\DeclareMathAlphabet\mathbfcal{OMS}{cmsy}{b}{n}
\usepackage{color,soul}
\usepackage{amsmath} % Mathe
\usepackage{mathtools} % Mathe
\usepackage{amsfonts} % Mathesymbole
\usepackage{amssymb}

\usepackage{array}
\usepackage{booktabs}
\usepackage{rotating}
\usepackage{multirow}
\usepackage{multicol}
\usepackage{lscape}
\usepackage{subfig}

\usepackage[export]{adjustbox}

\usepackage{algorithm,algorithmicx,algpseudocode}

\usepackage[labelfont=bf]{caption}
\usepackage[colorlinks=true,linkcolor=blue,citecolor=blue]{hyperref}

\usepackage{color}

\usepackage{xargs}                      % Use more than one optional parameter in a new commands

\usepackage[colorinlistoftodos,prependcaption,textsize=tiny]{todonotes}

\usepackage{tikz,xcolor,hyperref}

\definecolor{lime}{HTML}{A6CE39}
\DeclareRobustCommand{\orcidicon}{
	\begin{tikzpicture}
	\draw[lime, fill=lime] (0,0) 
	circle [radius=0.16] 
	node[white] {{\fontfamily{qag}\selectfont \tiny ID}};
	\draw[white, fill=white] (-0.0625,0.095) 
	circle [radius=0.007];
	\end{tikzpicture}
	\hspace{-2mm}
}

\foreach \x in {A, ..., Z}{\expandafter\xdef\csname orcid\x\endcsname{\noexpand\href{https://orcid.org/\csname orcidauthor\x\endcsname}
			{\noexpand\orcidicon}}
}
			
\definecolor{darkgreen}{rgb}{0.53, 0.66, 0.42}

\begin{document}

\title{Topology-Aware Generative Adversarial Network for Joint Prediction of Multiple Brain Graphs from a Single Brain Graph}

\titlerunning{Short Title}  % abbreviated title (for running head)

\author{Alaa Bessadok\index{Bessadok, Alaa}\inst{1,2}, Mohamed Ali Mahjoub\inst{2} \and Islem Rekik\index{Rekik, Islem}\orcidA{}\inst{1}\thanks{ {corresponding author: irekik@itu.edu.tr, \url{http://basira-lab.com}. This work is accepted for publication at MICCAI 2020.}} }

\authorrunning{A Bessadok et al.}  

\institute{$^{1}$ BASIRA Lab, Faculty of Computer and Informatics, Istanbul Technical University, Istanbul, Turkey \\ $^{2}$ Universit\'e de Sousse, Higher Institute of Informatics and Communication Technologies, Sousse, Tunisia}
\maketitle

\begin{abstract}
Multimodal medical datasets with incomplete observations present a barrier to large-scale neuroscience studies. Several works based on Generative Adversarial Networks (GAN) have been recently proposed to predict a set of medical images from a single modality (e.g, FLAIR MRI from T1 MRI). However, such frameworks are primarily designed to operate on \emph{images}, limiting their generalizability to non-Euclidean geometric data such as brain graphs. While a growing number of connectomic studies has demonstrated the promise of including brain graphs for diagnosing neurological disorders, no geometric deep learning work was designed for \emph{multiple target brain graphs prediction from a source brain graph.} Despite the momentum the field of graph generation has gained in the last two years, existing works have two critical drawbacks. \emph{First}, the bulk of such works aims to learn one model for each target domain to generate from a source domain. Thus, they have a limited scalability in \emph{jointly} predicting multiple target domains. \emph{Second}, they merely consider the global topological scale of a graph (i.e., graph connectivity structure) and overlook the local topology at the node scale of a graph (e.g., how central a node is in the graph). To meet these challenges, we introduce MultiGraphGAN architecture, which not only predicts multiple brain graphs from a single brain graph but also preserves the topological structure of each target graph to predict. Its three core contributions lie in: (i) designing a graph adversarial auto-encoder for jointly predicting brain graphs from a single one, (ii) handling the mode collapse problem of GAN by clustering the encoded source graphs and proposing a \emph{cluster-specific decoder}, (iii) introducing a \emph{topological loss} to force the reconstruction of topologically sound target brain graphs. Our MultiGraphGAN significantly outperformed its variants thereby showing its great potential in multi-view brain graph generation from a single graph. Our code is available at \url{https://github.com/basiralab/MultiGraphGAN}.
\end{abstract}


%% ***************************************************************************** %%
\section{Introduction}
%% ***************************************************************************** %%

Multimodal image synthesis has gained a lot of attention from researchers in the medical field as it reduces the high acquisition time and cost of medical modalities (e.g, positron emission tomography (PET)). Generative Adversarial Network (GAN) \cite{Goodfellow:2014} is nowadays the dominant method for predicting medical images of different modalities from a given modality. For instance, \cite{Pan:2019} proposed a GAN-based framework to predict PET neuroimaging from magnetic resonance imaging (MRI) for an early Alzheimer's disease diagnosis. Inspired from CollaGAN, \cite{Li:2019} predicted double inversion recovery (DIR) scans from three source modalities (i.e., Flair, T1 and T2). However, such \emph{one-target prediction} frameworks are incapable of \emph{jointly} predicting multiple target modalities using a single learning model. To alleviate this issue, several \emph{multi-target prediction} solutions have been proposed \cite{Wu:2019,Cao:2019} in the computer vision field but a few attempts have been made in the medical field. Recently, \cite{Huang:2019} proposed an adversarial autoencoder framework to predict three target MRI images (i.e., T1-weighted, T2-weighted, and FLAIR) from a single source T1 MRI scan. Although promising, such models fail to generalize to geometric data such as graphs and manifolds, especially brain graphs (i.e., connectome) which are derived from MRI scans. A brain graph consists of a set of nodes representing the anatomical regions of interest (ROIs) linked by edges encoding their biological relationship. However, multimodal medical datasets are usually incomplete so it becomes very challenging to conduct multimodal connectomic studies requiring paired samples. Consequently, predicting missing brain graphs from an existing source graph is highly desired since it provides rich and complementary information for brain mapping and disease diagnosis.

So far, we have identified only two brain graph synthesis works \cite{Bessadok:2019a,Bessadok:2019b} which proposed a geometric deep learning-based framework for \emph{one-target prediction} from a source brain graph. The target graph of a testing subject is predicted by first aligning the training target graphs to the source graphs, then averaging the target graphs of the training subjects that share similar local neighborhoods across source and target domains. Although pioneering, these works are neither designed in an \emph{end-to-end learning} manner nor effective for \emph{jointly} predicting multiple target brain graphs from a single source graph. Other works \cite{Su:2019,Liao:2019,Flam:2020,Bresson:2019} aimed to generate different types of graphs including biological ones such as molecules. To the best of our knowledge, no existing graph synthesis works attempted to solve the problem of \emph{joint} multiple brain graph prediction from a baseline source graph \cite{Zhang:2018,Zhou:2018}. Another important shortcoming of existing graph synthesis works \cite{Su:2019,Liao:2019,Flam:2020,Bresson:2019,Zhang:2018,Zhou:2018}, is that they do not preserve the node-wise topological properties. Mainly, they only learn the \emph{global graph structure} (i.e., number of nodes and edges weights). However, the brain wiring has both global and local topological properties underpinning its function, and which can get altered in neurological disorders \cite{Fornito:2015,Heuvel:2019}. Hence, by overlooking the learning of the \emph{local graph structure} one cannot capture which ROIs would be most effective for early diagnosing the disease based on the topological properties within the brain graph. By considering local topological constraints, one can learn the node's importance in a graph which can be measured using path-length based metrics such as betweenness centrality. Such centrality metrics assign a score to each node based on the shortest path between pairs of nodes. In this way, the synthesized graph will satisfy both global and local topologies of the original target graph.

\begin{figure}[htp!]
\centering
{\includegraphics[width=12cm]{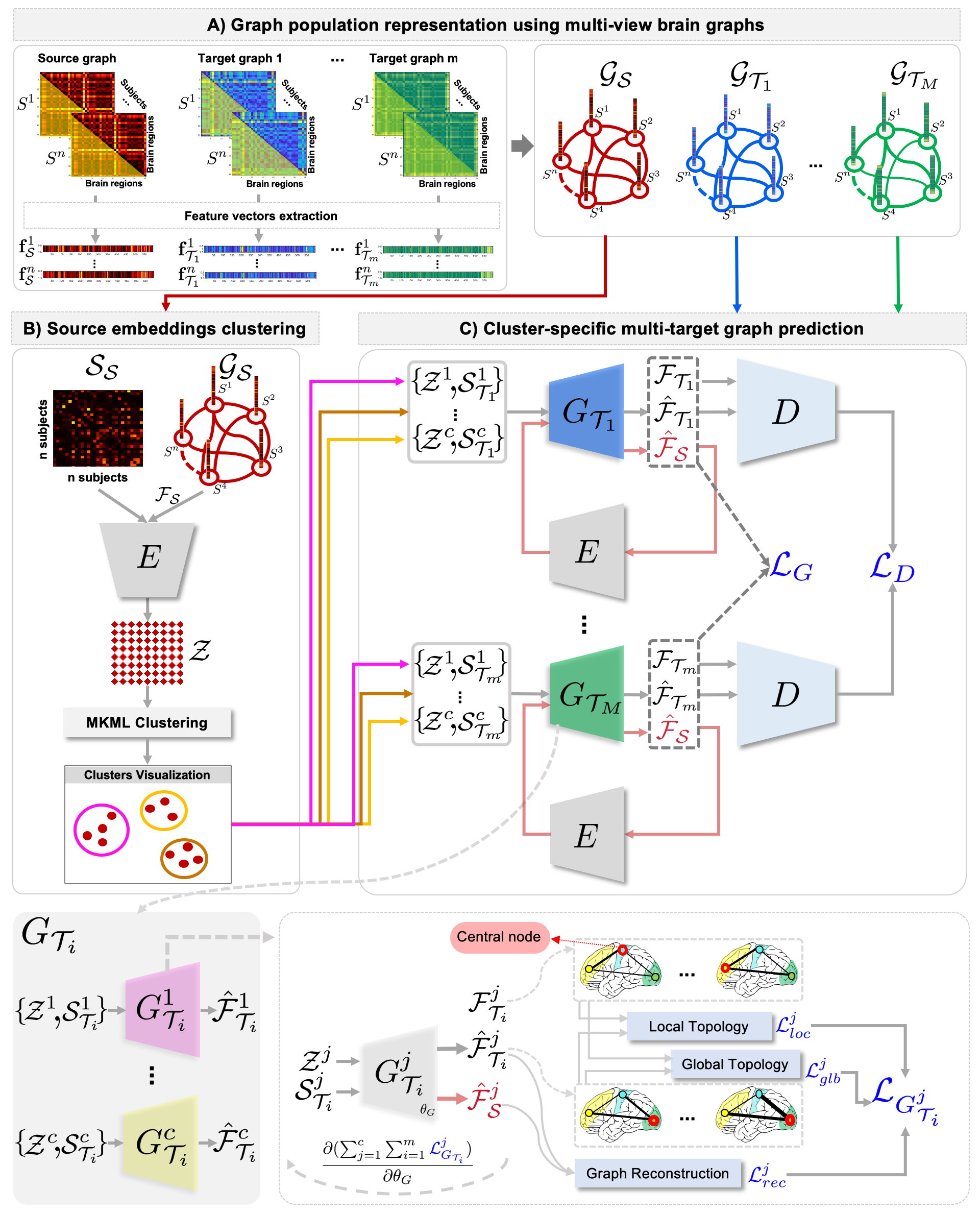}}
\caption{\emph{Pipeline of the proposed MultiGraphGAN framework for predicting jointly multiple target brain graphs from a single source graph.} \textbf{(A) Graph population representation using multi-view brain graphs.} Extraction of feature vectors from source and $m$ target brain graphs for each subject. Construction of graph population denoting the similarity between subjects using the resulting features. \textbf{(B) Source embeddings clustering.} First, we learn the source graph embedding using an encoder $E$. Second, we use multiple kernel manifold learning to cluster the resulting source embeddings into $c$ groups. \textbf{(C) Cluster-specific multi-target graph prediction.} For each of the $m$ target domains, we train $c$ \emph{cluster-specific generators} regularized by a shared discriminator $D$. We introduce a \emph{local topology loss} and a \emph{global topology loss} to regularize the cluster-specific generators (e.g., $G^{j}_{\mathcal{T}_{i}}$), each preserving the local node topology and the global graph connectivity structure. We further propose a source \emph{graph reconstruction loss} to map the generated target graphs back to the source domain.}
\label{fig:1}
\end{figure}

To address all these drawbacks, we propose MultiGraphGAN, the first attempt to \emph{jointly} predict multiple brain graphs from a single graph in an end-to-end deep learning fashion. We draw inspiration from the work \cite{Cao:2019} on multi-domain image translation task. Although effective for multi-target \emph{image} prediction, \cite{Cao:2019} fails to operate on graphs as it was primarily designed for Euclidean data. Besides, it overlooks GAN mode collapse, where the generator (i.e., decoder) produces data that mimic a few modes of the target domain. To address this issue, we first propose to learn the source graph embeddings using an encoder $E$ defined as a Graph Convolutional Network (GCN) \cite{Kipf:2016}. Second, we cluster the resulting embeddings with heterogeneous distribution into homogeneous clusters where a \emph{cluster-specific} generator is constructed to generate a specific mode of the given target domain. In other words, we define for each target domain a set of synergetic generators, each representing a cluster-specific GCN decoder. Hence, the graph prediction is learned more synergistically using our proposed cluster-specific generators, rather than using a single generator for each target domain. This generative process is regularized using one discriminator $D$, which enforces the generated graphs to match the original target graphs. Lastly, we introduce a \emph{topology-aware adversarial loss function} that seeks to preserve both global and local topological properties when predicting the target graphs. Mainly, we aim to enforce the generated graphs to retain a centrality score of each nodes in the original target brain graph.

% %% ***************************************************************************** %%
\section{Proposed MultiGraphGAN for Multiple Graphs Prediction}
% %% ***************************************************************************** %%

In the following, we present the main steps of our joint \emph{multi-target} brain graphs prediction framework from a single source graph. \textbf{Fig.}~\ref{fig:1} provides an overview of the key three steps of the proposed framework: 1) extraction of multi-view brain features and construction of a graph population for each source and target domains, 2) embedding and clustering of the source graphs, and 3) prediction of multiple target brain graphs using cluster-specific generators.

\textbf{A- Graph population representation using multi-view brain graphs}. Let $\mathbfcal{G}_{d}$ be a graph encoding the pairwise relationship between subjects belonging to a specific domain $d$ where $d\in\{\mathcal{S},\mathcal{T}_{1},\dots,\mathcal{T}_{m}\}$. We define our graph population as $\mathbfcal{G}_{d}=\{(\mathbfcal{G}^{n}_{d},\mathbfcal{F}_{d}), \mathbfcal{G}^{e}_{d}\}$ where $\mathbfcal{G}^{n}_{d}$ denotes a set of nodes (i.e., subjects) and  $\mathbfcal{F}_{d}$ denotes a feature matrix in $\mathbb{R}^{n\times f}$ vertically stacking the brain graph features of size $f$ for $n$ subjects. Specifically, each subject is represented by one source brain graph and $m$ target graphs where each graph is encoded in a symmetric matrix whose elements measure the similarity between two ROIs (i.e.,nodes). We vectorize the off-diagonal upper-diagonal part of each matrix to create a feature vector ${\mathbf{f}}_{d}$ in $\mathbb{R}^{1\times f}$ encoding the connectivity features of a subject in the domain $d$. Thus, $\mathbfcal{F}_{d}$ denotes the feature vectors $\{\mathbf{f}_{d}^{1},\dots,\mathbf{f}_{d}^{n}\}$ of $n$ subjects. Additionally, we define $\mathbfcal{G}^{e}_{d}$ as a set of weighted edges encoding the similarity between each pair of subjects using their feature vectors. To do this, we propose to learn a sample similarity matrix $\mathbfcal{S}_{d}$ in $\mathbb{R}^{n\times n}$ using multi-kernel manifold learning (MKML) algorithm \cite{Wang:2017} as it efficiently fits the statistical distribution of the data by learning multiple kernels. Ultimately, for the source and $m$ target domains we have a set of graphs $\{\mathbfcal{G}_{\mathcal{S}}, \mathbfcal{G}_{\mathcal{T}_{1}}, \dots ,\mathbfcal{G}_{\mathcal{T}_{m}} \}$ each represented by a set of feature matrices $\{\mathbfcal{F}_{\mathcal{S}}, \mathbfcal{F}_{\mathcal{T}_{1}}, \dots , \mathbfcal{F}_{\mathcal{T}_{m}} \}$ and a set of learned adjacency matrices $\{\mathbfcal{S}_{\mathcal{S}}, \mathbfcal{S}_{\mathcal{T}_{1}}, \dots, \mathbfcal{S}_{\mathcal{T}_{m}} \}$ (\textbf{Fig.}~\ref{fig:1}--A).
 
\textbf{B- Source graphs embedding and clustering.} We aim in this step to learn the source graph embeddings using an encoder $E(\mathbfcal{F}_{\mathcal{S}},\mathbfcal{S}_{\mathcal{S}})$ defined as a GCN with two layers inputing the source feature matrix $\mathbfcal{F}_{\mathcal{S}}$ and the learned sample similarity matrix $\mathbfcal{S}_{\mathcal{S}}$. We define the layers of GCN and the graph convolution function used in each layer as follows:

\begin{gather}
	\mathbfcal{Z}^{(l)} = f_{\phi}(\mathbfcal{X}, \mathbfcal{S}_{\mathcal{S}} \vert \mathbf{W}^{(l)}); 
    	\quad\text{ }
    	f_{\phi}(\mathbfcal{X}^{(l)}, \mathbfcal{S}_{\mathcal{S}} \vert \mathbf{W}^{(l)}) = \phi(\mathbf{\widetilde{D}}^{-\frac{1}{2}} \mathbf{\widetilde{\mathbfcal{S}_{\mathcal{S}}}}\mathbf{\widetilde{D}}^{-\frac{1}{2}}\mathbfcal{X}^{(l)}\mathbf{W}^{(l)}),
\label{eq:1}
\end{gather}

$\mathbfcal{Z}^{(l)}$ is the resulting source graph embeddings of the layer $l$. $\phi$ represents the $ReLU$ and $linear$ activation functions we used in the first and second layers, respectively. In the first layer, $\mathbfcal{X}$ denotes the source feature matrix $\mathbfcal{F}_{\mathcal{S}}$ while in the second layer it denotes the resulting embeddings learned from the first layer $\mathbfcal{Z}^{(1)}$. $\mathbf{W}^{(l)}$ is a filter used to learn the convolution in the GCN in each layer $l$. As in \cite{Kipf:2016}, we define the graph convolution function by $f_{(.)}$ where $\mathbf{\widetilde{\mathbfcal{S}_{\mathcal{S}}}} = \mathbf{\mathbfcal{S}_{\mathcal{S}}} + \mathbf{I}$ with $\mathbf{I}$ being an identity matrix used for self-regularization, and $\mathbf{\widetilde{D}}_{ii} = \sum_{j}\mathbf{\widetilde{\mathbfcal{S}_{\mathcal{S}}}}(ij)$ is a diagonal matrix.

We aim in the following step to build a set of domain-specific decoders regularized with the discriminator $D$ to generate the target graphs. However, in practice, the GAN generators might end up producing graphs that match a few unimodal sample of the target domain thereby overlooking its heterogeneous distribution. To handle such mode collapse of generative models, we propose to first cluster the source graph embeddings $\mathbfcal{Z}$ into homogeneous clusters. We further use MKML for clustering since it outperformed PCA and t-SNE clustering methods when dealing with biological datasets \cite{Wang:2017}. Specifically, it first learns the similarity between source embeddings, second it maps the learned similarity matrix into a lower dimensional space, and finally uses k-means algorithm to cluster the subjects into $c$ clusters (\textbf{Fig.}~\ref{fig:1}--B).

\textbf{C- Cluster-specific multi-target graph prediction.} To predict the target graph of a given domain ${\mathcal{T}_{i}}$ where $i \in \ \{1, \dots, m \}$, we propose a set of \emph{cluster-specific generators} ${G_{\mathcal{T}_{i}}} = \{{G_{\mathcal{T}_{i}}^{1}}, \dots , {G_{\mathcal{T}_{i}}^{c}} \}$, where each generator produces a graph approximating the target data distribution of a specific cluster (\textbf{Fig.}~\ref{fig:1}--C). As such, we enforce the generator to learn from all examples in the cluster $c$ thus avoiding the mode collapse issue as our learning becomes unimodal (i.e., cluster-specific). We train the generators in a sequential manner where each is defined as a GCN decoder with a similar architecture to the encoder (\textbf{Eq.}~\eqref{eq:1}). More specifically, for each cluster $j$, a generator ${G}_{\mathcal{T}_{i}}^{j}$ assigned to the target domain $\mathcal{T}_{i}$ and to the cluster $j$ takes two inputs: the source embeddings $\mathbfcal{Z}^{j}$ and the sample similarity matrix $\mathbfcal{S}_{\mathcal{T}_{i}}^{j}$ learned using the target graphs in domain $\mathcal{T}_{i}$. In that way, we enforce the generator to decode the source embeddings while approximating the real target graph structure. 

The target graph prediction is optimized using the discriminator $D$ which is a GCN with three layers. Specifically, it enforces the generated target graph to match the ground truth target distribution of a specific target domain. This is achieved in two steps. First, the discriminator measures the realness of the generated graphs by computing the Wasserstein distance among all domains. We formulate this using the following adversarial loss $\mathcal{L}_{adv}^{j} = - \mathbb{E}_{\mathbfcal{F}^{\prime} \sim \mathbb{P}_{{\mathbfcal{F}}^{j}_{\mathcal{S}}}} \ [ D(\mathbfcal{F}^{\prime}) \ ] + \frac{1}{m} \sum_{i=1}^{m} \mathbb{E}_{\mathbfcal{F}^{\prime\prime} \sim \mathbb{P}_{{\hat{\mathbfcal{F}}}^{j}_{\mathcal{T}_{i}}} }\ [ D(\mathbfcal{F}^{\prime\prime}) \ ]$. Second, we define a binary classifier $D_{C}$ on top of our discriminator $D$ which classifies the fake graphs $\hat{\mathbfcal{F}}^{j}_{\mathcal{T}_{i}}$ as $0$ and the real target graphs $\mathbfcal{F}^{j}_{\mathcal{T}_{i}}$ as $1$. Hence, we formulate a graph domain classification loss as $\mathcal{L}_{gdc}^{j} = \sum_{i=1}^{m} \mathbb{E}_{\mathbfcal{F}^{\prime\prime}\sim \mathbb{P}_{\hat{\mathbfcal{F}}^{j}_{\mathcal{T}_{i}}} \cup  \mathbb{P}_{{\mathbfcal{F}}^{j}_{\mathcal{T}_{i}}}} \ [ \ell_{MSE}(D_{C}(\mathbfcal{F}^{\prime\prime}),y({\mathbfcal{F}^{\prime\prime}})) \ ]$. $\ell_{MSE}$ is the mean squared loss and $y$ is the ground truth label corresponding to the graph $\mathbfcal{F}^{\prime\prime}$. Additionally, to improve the training stability of our model we adopt the gradient penalty loss of \cite{Cao:2019} which is formulated as $\mathcal{L}_{gp}^{j} = (max \{ 0, {\mathbb{E}_{\tilde{\mathbfcal{F}}\sim{\mathbb{P}_{\tilde{{\mathbfcal{F}}}^{j}_{m}}}}} \vert\vert \nabla  D(\tilde{\mathbfcal{F}}) \vert\vert - \sigma  \} )^{2}$. $\tilde{\mathbfcal{F}}$ is sampled between the source graph distribution $\mathbb{P}_{{\mathbfcal{F}}^{j}_{\mathcal{S}}}$ and the fake target graph distribution $\mathbb{P}_{\tilde{\mathbfcal{F}}^{j}_{m}}$ where ${\tilde{\mathbfcal{F}}^{j}_{m}}$ is a matrix stacking vertically the generated target graphs for all $m$ domains. In particular, $\tilde{\mathbfcal{F}} \leftarrow \alpha{\mathbfcal{F}}^{j}_{\mathcal{S}} + (1-\alpha)\tilde{\mathbfcal{F}}^{j}_{m}$ where $\alpha \sim U\ [0,1\ ]$ and $U$ is a uniform distribution. $\sigma$ is a hyper-parameter set to $m$ as suggested in \cite{Cao:2019}. Ultimately, the discriminator guides the generators of each cluster to produce brain graphs, each associated with a specific target domain through the following loss function:

\begin{equation}
 	\mathcal{L}_{D} = \sum_{j=1}^{c} ( \mathcal{L}_{adv}^{j} + \lambda_{gdc} \cdot \mathcal{L}_{gdc}^{j}+ \lambda_{gp} \cdot \mathcal{L}_{gp}^{j} ),
\label{eq:2}
\end{equation}

$\lambda_{gdc}$ and $\lambda_{gp}$ are hyper-parameters to be tuned. Moreover, brain graphs have unique topological properties for functional, structural and morphological connectivities that should be preserved when predicting the target brain graphs \cite{Liu:2017,Joyce:2010}. To this aim, we introduce a topological loss function which constrains the generators to preserve the nodes properties while learning the global graph structure (\textbf{Fig.}~\ref{fig:1}--C). To do so, we compute the absolute difference between the real and predicted centralities scores of each node in the target graph. We choose three centrality measures widely used in graph theory: closeness centrality $CC$ quantifying the closeness of a node to all other nodes \cite{Freeman:1977}, betweenness centrality $BC$ measuring the number of shortest paths which pass across a node \cite{Beauchamp:1965}, and eigenvector centrality $EC$ capturing the centralities of a node's neighbors \cite{Bonacich:2007}. We define their formulas in \textbf{Table.}~\ref{tab:0}.

\begin{table*}[h]
\centering
\scalebox{0.8}{
\begin{tabular}{| c | c |} 
\hline
Centrality  & Description \\ \hline 
$CC(v^{a}) = \frac{V-1}{\sum_{v^{a}\neq v^{b}} p_{v^{a}v^{b}}}$ &
$V$ is the number of nodes and $p_{v^{a}v^{b}}$ is the length of \\& the shortest path between nodes $v^{a}$ and $v^{b}$ \\ \hline
$BC(v^{a}) = {\frac{2}{(V-1)(V-2)}} \times {\sum_{v^{a} \neq v^{b} \neq v^{c}}} \frac{P_{(v^{c},v^{b})} (v^{a})}{P_{(v^{c},v^{b})}}$ &
$P_{(v^{c},v^{b})} (v^{a})$ denotes the number of shortest paths \\ & between two nodes $v^{c}$ and $v^{b}$ that pass through $(v^{a})$ \\ \hline 
$EC(v^{a}) = x^{a} = \frac{1}{\lambda} \sum_{h=1}^{V}  {A}_{a b} {x}^{b}$ &
${A}_{a b}$ represents all neighbors of the node $a$, $x$ is the \\ & eigenvector resulted from the eigen decomposition \\ & of the adjacency matrix $A$ and $\lambda$ is the highest eigen value \\ \hline
\end{tabular}}
\caption{\emph{Centrality measures included in the topological loss function.}}
\label{tab:0}
\end{table*}

Given a centrality metric $\mathcal{C}$ where $\mathcal{C}\in\{CC, BC, EC\}$, a cluster $j$ and a target domain $\mathcal{T}_{i}$, we define $\mathbfcal{X}_{\mathcal{T}_{i}}^{j}$ and $\hat{\mathbfcal{X}}_{\mathcal{T}_{i}}^{j}$ as the centralities for the real graphs $\mathbfcal{F}_{\mathcal{T}_{i}}^{j}$ and the generated ones $\hat{\mathbfcal{F}}_{\mathcal{T}_{i}}^{j}$, respectively. Both $\mathbfcal{X}$ matrices are in $\mathbb{R}^{n\times r}$ where $n$ is the number of subjects and $r$ is the number of brain regions. Hence, we define our proposed \emph{local topology} loss as $\mathcal{L}^{j}_{loc} (\mathcal{C}) = \sum_{i=1}^{m} \ell_{MAE}(\mathbfcal{X}_{\mathcal{T}_{i}}^{j}, \hat{\mathbfcal{X}}_{\mathcal{T}_{i}}^{j})$. On the other hand, we propose the \emph{global topology} loss function to maintain the relationship between brain regions in terms of number of edges and their weights using the feature matrix $\mathbfcal{F}^{j}_{\mathcal{T}_{i}}$. Hence, for a cluster $j$, we define it as $\mathcal{L}^{j}_{glb} = \sum_{i=1}^{m} \ell_{MAE}(\mathbfcal{F}_{\mathcal{T}_{i}}^{j}, \hat{\mathbfcal{F}}_{\mathcal{T}_{i}}^{j})$. One of the key contributions for our proposed architecture is the topological loss function regularizing the cluster-specific generators. It is made up of local and global topology losses and defined as $\mathcal{L}^{j}_{top} = \mathcal{L}_{loc}^{j} + \mathcal{L}_{glb}^{j}$. Moreover, by maximizing the \textbf{Eq.}~\eqref{eq:2} the generators are optimally trained to produce graphs that belong to a specific target domain. However, this does not guarantee that the predicted target graphs can inversely regenerate the source graph structure in a cyclic manner. To address this problem, we propose a graph reconstruction loss function which ensures that the source brain graphs can be also generated from the predicted brain graphs (\textbf{Fig.}~\ref{fig:1}--C). Similar to the topological loss function $\mathcal{L}^{j}_{top}$, we define it as follows: 

\begin{equation}
 	\mathcal{L}^{j}_{rec} = (\underbrace{\sum_{i=1}^{m} \ell_{MAE}(\mathbfcal{X}_{\mathcal{S}_{i}}^{j}, \hat{\mathbfcal{X}}_{\mathcal{S}_{i}}^{j})}_{\text{ reconstruction local topology loss}} + \underbrace{\sum_{i=1}^{m} \ell_{MAE}(\mathbfcal{F}_{\mathcal{S}_{i}}^{j}, \hat{\mathbfcal{F}}_{\mathcal{S}_{i}}^{j})}_{\text{reconstruction global topology loss}} )
\label{eq:9}
\end{equation}
Furthermore, since the target domains are correlated we integrate the information maximization loss term to force the generators to correlate the predicted graphs with a specific target domain. It is defined as in \cite{Cao:2019} $\mathcal{L}_{inf}^{j} = \sum_{i=1}^{m} \ell_{BCE}(y=1,D_{C}(\hat{\mathbfcal{F}}_{\mathcal{T}_{i}}^{j}))$ where $\ell_{BCE}$ is the binary cross entropy. Ultimately, in our MultiGraphGAN architecture, we define the overall \emph{topology-aware adversarial loss function} of each generator as:

\begin{equation}
 	\mathcal{L}_{G} = \sum_{j=1}^{c} ( - \frac{1}{m} \cdot \sum_{i=1}^{m} \mathbb{E}_{\mathbfcal{F}^{\prime\prime} \sim {\hat{\mathbfcal{F}}}_{\mathcal{T}_{i}}} \ [ D(\mathbfcal{F}^{\prime\prime}) \ ] + \lambda_{top} \cdot \mathcal{L}^{j}_{top} + \lambda_{rec} \cdot \mathcal{L}^{j}_{rec} + \lambda_{inf} \cdot \mathcal{L}^{j}_{inf}),
\label{eq:11}
\end{equation}

where $\lambda_{top}$, $\lambda_{rec}$ and $\lambda_{inf}$ are hyper-parameters that control the relative importance of topological loss, graph reconstruction, and information maximization losses, respectively. The steps explained above are used for training our MultiGraphGAN and for a testing subject we predict its target graph by averaging the target graphs produced by the cluster-specific generators.

% %% ***************************************************************************** %%
\section{Results and Discussion}
% %% ***************************************************************************** %%

\textbf{Multi-view brain graph dataset and model architecture.} A set of 310 structural T1-w MRI data extracted from Autism Brain Imaging Data Exchange (ABIDE\footnote{\url{http://fcon\_1000.projects.nitrc.org/indi/abide/}})  public dataset was used. We train our model on 90\% of the dataset and test it on 10\%. Each subject is represented by six morphological brain graphs (MBG). For each hemisphere $H$ (i.e., $H\in\{L,R\}$), we extract three MBGs using the following cortical measurements as introduced in \cite{Mahjoub:2018}: $MBG^{1}_{H}$ maximum principal curvature, $MBG^{2}_{H}$ average curvature and $MBG^{3}_{H}$ mean sulcal depth. We consider $MBG^{1}_{L}$ as the source brain graphs and $\{MBG^{2}_{L}, MBG^{3}_{L}, MBG^{1}_{R}, MBG^{2}_{R}, MBG^{3}_{R}\}$ as the target graphs. We construct our encoder with a hidden layer comprising 32 neurons and an embedding layer with 16 neurons. Conversely, we define all generators with two layers each comprising 16 and 32 neurons. The discriminator comprises three layers each has 32, 16 and 1 neurons, respectively. We add to its last layer a softmax activation function representing our domain classifier. We train our model using 1000 iterations, a batch size of 70, a learning rate of 0.0001, $\beta_{1} = 0.5$ and $\beta{2} = 0.999$ for Adam optimizer. Using grid search we set our hyper-parameters $\lambda_{gdc}=1$, $\lambda_{gp}=0.1$, $\lambda_{top}=0.1$, $\lambda_{rec}=0.01$ and $\lambda_{inf}=1$. We train the discriminator five times and the generators one time in an iterative manner so that their learning performances are improved. For MKML parameters \cite{Wang:2017}, we fix the number of kernels to 10. After evaluating our model on different number of clusters $c\in\{2,3,4\}$ we choose the one which gave the best performance $c=2$.

\begin{table*}[h]
\centering
\scalebox{0.93}{
\begin{tabular}{c c c c c c}
\hline
{\color[HTML]{333333} Methods}                     & {\color[HTML]{333333} Topological measures} & {\color[HTML]{333333} PCC}    & {\color[HTML]{333333} MAE (BC)} & {\color[HTML]{333333} MAE (CC)} & {\color[HTML]{333333} MAE (EC)} \\ \hline
{\color[HTML]{333333} Adapted MWGAN \cite{Cao:2019} }              & {\color[HTML]{333333} \_\_}                 & {\color[HTML]{333333} 0.4869} & {\color[HTML]{333333} 0.0101} & {\color[HTML]{333333} 0.2394} & {\color[HTML]{333333} 0.0169} \\ 
{\color[HTML]{333333} Adapted MWGAN \cite{Cao:2019} (clustering)} & {\color[HTML]{333333} \_\_}                 & {\color[HTML]{333333} 0.4272} & {\color[HTML]{333333} 0.0063} & {\color[HTML]{333333} 0.1624} & {\color[HTML]{333333} 0.013}  \\ \hline \hline
{\color[HTML]{333333} MultiGraphGAN}        & {\color[HTML]{333333} CC}                   & {\color[HTML]{333333} 0.3428} & {\color[HTML]{333333} 0.0062} & {\color[HTML]{333333} 0.1599} & {\color[HTML]{333333} 0.0118} \\ 
{\color[HTML]{333333} \textbf{MultiGraphGAN}}        & {\color[HTML]{333333} \textbf{BC}}                   & {\color[HTML]{333333} \textbf{0.5037}} & {\color[HTML]{3166FF} \textbf{0.0054}} & {\color[HTML]{3166FF} \textbf{0.141}}  & {\color[HTML]{333333} \textbf{0.0113}} \\ 
{\color[HTML]{333333} \textbf{MultiGraphGAN}}       & {\color[HTML]{333333} \textbf{EC}}                   & {\color[HTML]{3166FF} \textbf{0.5245}} & {\color[HTML]{333333} \textbf{0.0056}} & {\color[HTML]{333333} \textbf{0.1449}} & {\color[HTML]{3166FF} \textbf{0.0111}} \\ \hline
\end{tabular}}
\caption{\emph{Prediction results using different evaluation metrics. PCC: pearson correlation coefficient. MAE: mean absolute error. BC: betweenness centrality. CC: closeness centrality. EC: eigenvector centrality.} }
\label{tab:1}
\end{table*}

\textbf{Evaluation and comparison methods.} As our MultiGraphGAN is the first model aiming to jointly predict multiple target graphs from a single brain graph, we compare it with two baseline methods: \textbf{(1) Adapted MWGAN:} we use the same architecture proposed in \cite{Cao:2019} that we adapted to graph data types where we neither include the clustering step nor our proposed topology-aware loss function. \textbf{(2) Adapted MWGAN (clustering):} it is a variant of the first method where we add the MKML clustering of the source graph embeddings \cite{Wang:2017} without any topology loss. We also compare our model when using three different centrality metrics: closeness \textbf{(3) MultiGraphGAN+CC}, betweenness \textbf{(4) MultiGraphGAN+BC} and eigenvector \textbf{(5) MultiGraphGAN+EC} centralities. To evaluate our framework, we compute for each target domain the Pearson Correlation Coefficient (PCC) between the ground truth and predicted graphs and the mean absolute error (MAE) between the ground truth centrality scores and the predicted ones. Then, we consider the average of all resulting PCCs, and MAEs as the final measures to evaluate our framework. \textbf{Table.}~\ref{tab:1} shows the outperformance of MultiGraphGAN over the baseline methods, which demonstrates the advantage of our \emph{cluster-specific generators} in avoiding the mode collapse problem in addition to the \emph{topological constraint} in optimally learning the target graph structure. Notably, the results also highlight the importance of using BC and EC which both ranked first best and second best using different evaluation metrics. This is explicable since considering the node neighborhoods (i.e., EC) and the frequency of being on the shortest path between nodes in the graph (i.e., BC) have much impact on identifying the most influential node rather than focusing on the average shortest path existing between a pair of nodes. As our MultiGraphGAN achieved very promising results, it can be extended in different directions such as predicting multi-target time-dependent brain graphs. This would be of high interest in foreseeing brain disorder evolution over time using brain graph representations \cite{Ezzine:2019,Ghribi:2019,Vohryzek:2020}.

% %% ***************************************************************************** %%
\section{Conclusion}
% %% ***************************************************************************** %%

We proposed MultiGraphGAN the first geometric deep learning framework for jointly predicting multiple target brain graphs from a single source graph. Our architecture has two compelling strengths: (i) clustering the learned source graph embeddings then training a set of \emph{cluster-specific generators} which synergistically predict the target brain graphs, (ii) introducing a \emph{topological loss function} using a centrality metric which enforces the generators to preserve local and global topology of the original target graphs. Our framework can be used for predicting other types of brain graphs such as structural and functional and extended to predict the evolution multi-target brain graphs over time from a single source brain graph.

% %% ***************************************************************************** %%
\section{Supplementary material}
% %% ***************************************************************************** %%

In addition to our open source code\footnote{\url{https://github.com/basiralab/MultiGraphGAN}}, we further provide three supplementary items on MultiGraphGan for reproducible and open science:

\begin{enumerate}
	\item A 5-mn YouTube video explaining how MultiGraphGAN works on BASIRA YouTube channel at \url{https://youtu.be/vEnzMQqbdHc}.
	\item A 20-mn detailed YouTube video on our work at \url{https://youtu.be/yNx7H9NLzlE}.
	\item A GitHub video code demo on BASIRA YouTube channel at \url{https://youtu.be/JvT5XtAgbUk}. 
\end{enumerate}

% %% ***************************************************************************** %%
\section{Acknowledgement}
% %% ***************************************************************************** %%

This project has been funded by the 2232 International Fellowship for
Outstanding Researchers Program of TUBITAK (Project No:118C288, \url{http://basira-lab.com/reprime/}) supporting I. Rekik. However, all scientific contributions made in this project
are owned and approved solely by the authors.

%%%%%%%%%%%%%%%%%%%%%%%%%%%%%%%%%%%%%%%%%%%%%%%%%%%%%%%%%%%%%%%%%%%%%%%%%%%%%%%%%%%%%%%%%%%%%%%%%%%%%%%%%%%%

\bibliography{Biblio3}
\bibliographystyle{splncs}
\end{document}